\documentclass[letterpaper, 10 pt, conference]{ieeeconf}
\IEEEoverridecommandlockouts%
\usepackage{graphics} %
\usepackage[tight]{subfigure}
\usepackage[pdftex]{graphicx}
\usepackage{amsfonts} %
\usepackage{amsmath} %
\usepackage{epstopdf}
\usepackage{multirow}
\usepackage[pdfpagemode=UseNone,pdftoolbar=false,pdfauthor=Alexander W Winkler / Carlos Mastalli /
Ioannis Havoutis,pdfstartview =FitH]{hyperref}
\usepackage{color,hyperref}
\definecolor{darkblue}{rgb}{0.0,0.0,0.0}
\hypersetup{colorlinks,breaklinks,
    linkcolor=black,urlcolor=darkblue,
    anchorcolor=black,citecolor=darkblue}   
\usepackage{color}
\usepackage[usenames,dvipsnames]{xcolor}
\usepackage{bm}
\usepackage[normalem]{ulem}	%

 \usepackage[leftcaption]{sidecap} %

\newcommand\eat[1]{}

\newcommand{\sref}[1]{Section~\ref{#1}}
\newcommand{\fref}[1]{Fig.~\ref{#1}}

\newcommand{\tref}[1]{Table~\ref{#1}}

\usepackage{booktabs}
\usepackage{vector}  %
\usepackage{glossaries}
	\newacronym{hyq}{HyQ}{Hydraulically Actuated Quadruped}
	\newacronym{cog}{CoG}{Center of Gravity}
  \newacronym{zmp}{ZMP}{Zero Moment Point}
  \newacronym{dof}{DoF}{Degree of Freedom}
\usepackage[tight]{units} %

\newcommand{\Rnum}{\mathbb{R}} %

\newcommand{\mx}[1]{\mathbf{\bm{#1}}} 				%
\newcommand{\vc}[1]{\mathbf{\bm{#1}}} 					%
\newcommand{\mat}[1]{\ensuremath{\begin{bmatrix}#1\end{bmatrix}}}	%

\newcommand{\dx}[1]{\ensuremath{\delta x_{#1}}}					%
\newcommand{\T}[0]{\ensuremath{\top}}							%

\begin{document}
\title{\LARGE \bf
Planning and Execution of Dynamic Whole-Body Locomotion\\
for a Hydraulic Quadruped on Challenging Terrain
}
\author{{\centering 
Alexander W. Winkler$^{\dagger *}$, \quad Carlos Mastalli$^*$, \quad Ioannis Havoutis$^*$,}\\\quad Michele Focchi$^*$, \quad Darwin G. Caldwell$^*$, \quad Claudio Semini$^*$
\thanks{
\hspace{-1em}$^*$Department of Advanced Robotics, Istituto Italiano di Tecnologia, 
via Morego, 30, 16163 Genova, Italy. \textit{email}: \{carlos.mastalli, ioannis.havoutis, 
michele.focchi, darwin.caldwell, claudio.semini\}@iit.it. 
\newline
$^\dagger$Agile \& Dexterous Robotics Lab, Institute of Robotics and Intelligent Systems, ETH Zurich, Switzerland. 
\textit{email}: alexander.winkler@mavt.ethz.ch
} 
}

\maketitle
\thispagestyle{empty}
\pagestyle{empty}

\begin{abstract}
We present a framework for dynamic quadrupedal locomotion over 
challenging terrain, where the choice of appropriate footholds is crucial for 
the success of the behaviour. We build a model of the environment on-line and 
on-board using an efficient occupancy grid representation. 
We use Any-time-Repairing A* (ARA*) to search over a tree of 
possible actions, choose a rough body path and select the locally-best
footholds accordingly.
We run a n-step lookahead optimization of the body trajectory using
a dynamic stability metric, the \gls{zmp}, that generates 
natural dynamic whole-body motions. A combination of 
floating-base inverse dynamics and virtual model control accurately executes
the desired motions on an actively compliant
system. Experimental trials show that this framework allows us to 
traverse terrains at nearly 6 times the speed of our previous work, 
evaluated over the same set of trials.  
\end{abstract}

\section{Introduction}\label{sec:introduction}
Agile locomotion is one of the key abilities that legged ground robots need to
master. Wheeled or tracked vehicles are efficient in structured
environments but can suffer from limited mobility in many real-world scenarios.
Legged robots offer a clear advantage in unstructured and 
challenging terrain. Such environments are common in disaster relief, search \& 
rescue, forestry and construction site scenarios. 

This paper presents the newest development in a stream of research that aims to 
increase the autonomy and flexibility of legged robots in unstructured and 
challenging environments. We present a framework for dynamic quadrupedal 
locomotion over highly challenging terrain where the choice of appropriate footholds is 
crucial for the success of the behaviour. We use perception to build a 
map of the environment, decide on a rough body path and choose appropriate 
footholds. We are able to generate feasible footholds \emph{on-line} 
and \emph{on-board} for various types of scenarios such as climbing up and 
down pallets, traversing stepping stones using an irregular swing-leg sequence 
and passing over a \unit[35]{cm} gap. 
We optimize the body
trajectory according to a dynamic stability metric (\gls{zmp}) to produce agile
and natural dynamic whole-body motions up to $5.8$ times the speed of our 
previous work \cite{winkler2014}. Compliant execution of 
the motions is performed using a floating-base inverse dynamics
controller that
ensures the accurate execution of dynamic motions, in combination with a
virtual model controller that generates feedback torques to account 
for model and tracking inaccuracies.
\begin{figure}[]
	\centering	
\subfigure{\resizebox{0.96\columnwidth}{!}{
\includegraphics{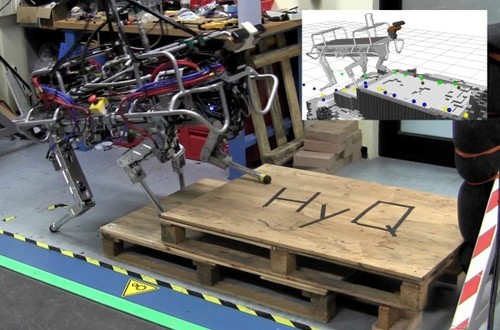}}}
	\caption{The hydraulically actuated and fully torque controlled quadruped
	robot HyQ \cite{Semini01092011}. The inset plot shows the
	on-line built environment perception alongside with the planned footholds
	and the current on-board robot state estimate.}
	\label{fig:hyq}
\end{figure}

Our contribution includes on-line perception, map building and foothold planning, generation and execution of optimized dynamic whole-body motions despite irregular swing-leg sequences and the use of an elegant inverse-dynamics/virtual-model control formulation that exploits the natural partitioning of the robot's dynamic equations.

The rest of the paper is structured as follows: After discussing previous research in the field of dynamic whole-body locomotion (\ref{sec:related_work}) we describe the on-line map building and how appropriate footholds are chosen (\ref{sec:planning}). \sref{sec:trajectory_generation} explains, how dynamically stable whole-body motions are generated based on an arbitrary footstep sequence. \sref{sec:execution} shows how these desired motions are accurately and compliantly executed. In \sref{sec:experimental_results} we evaluate the performance of our framework 
on the Hydraulic Quadruped robot \textit{HyQ} (\fref{fig:hyq})
in real-world experimental trials before \sref{sec:conclusion} summarizes this work and presents ideas for future work.

\section{Related Work}\label{sec:related_work}
In environments where smooth, continuous support is available (flats, fields,
roads, etc.), where exact foot placement is not crucial for the
success of the behaviour, legged systems can utilize a variety
of dynamic gaits, e.g. trotting, galloping. Marc Raibert pioneered the study of the principles of dynamic balancing with legged robots \cite{1986_Raibert}, resulting in the quadruped \emph{BigDog}. The  \emph{reactive} controllers used in these legged systems are partially capable to overcome unstructured terrain. Likewise, \emph{HyQ} can traverse lightly unstructured terrain using reactive control \cite{2013_Barasuol, havoutisICM13} or reflex strategies \cite{2013_focchi}.

However, for more complex environments with obstacles like large gaps or stairs,
such systems quickly reach their limits. In this case, higher level motion 
planning that considers the environment and carefully selects appropriate 
footholds is required. 
In these terrains, e.g. stairs, gaps, 
cluttered rooms, legged robots have the potential to use non-gaited 
locomotion strategies that rely more on accurate 
foothold planning based on features of the terrain.   
There exist a number of successful control 
architectures \cite{kalakrishnan2010IJRR,2007_Rebula,2008_Kolter,2011_Zucker}
to plan and execute footsteps to traverse 
such terrain. Some avoid global footstep 
planning by simply choosing the next best reachable foothold \cite{2007_Rebula},
while others plan the complete footstep sequence from start to goal \cite{2011_Zucker}, often 
requiring time consuming re-planning in case of slippage or deviation 
from the planned path. 

The approach in \cite{kalakrishnan2010IJRR} stands between the two above 
mentioned methods and plans a 
global rough body path to avoid local minima, but the specific footholds are 
chosen only a few steps in advance. This reduces the necessary time for 
re-planning in case of slippage, while still considering a locally optimal plan.
We recently built on this approach with a path planning and control framework that uses on-line 
force-based foothold 
adaptation to update the planned motion according to the perceived state of the 
environment during execution \cite{winkler2014}.

The whole-body locomotion framework described in this paper further extends this work: We use real-time perception to create, evaluate and update a terrain cost map \emph{on-board}. Compared to previous approaches our framework does not make use of any external state measuring system, e.g. a marker-based tracking system. The incorporation of domain knowledge, e.g. body motion primitives and an ARA* planner, allows us to re-plan actions and footholds \emph{on-line}.  
As in \cite{kalakrishnan2010IJRR} the \gls{cog} trajectory is now chosen to comply with the \gls{zmp} \emph{dynamic} stability metric \cite{Kajita2003} to produce agile, fast and natural motions.

\section{Perception and (Re-)Planning}\label{sec:planning}
\begin{figure}%
	\centering
	\includegraphics[width=0.7\columnwidth]{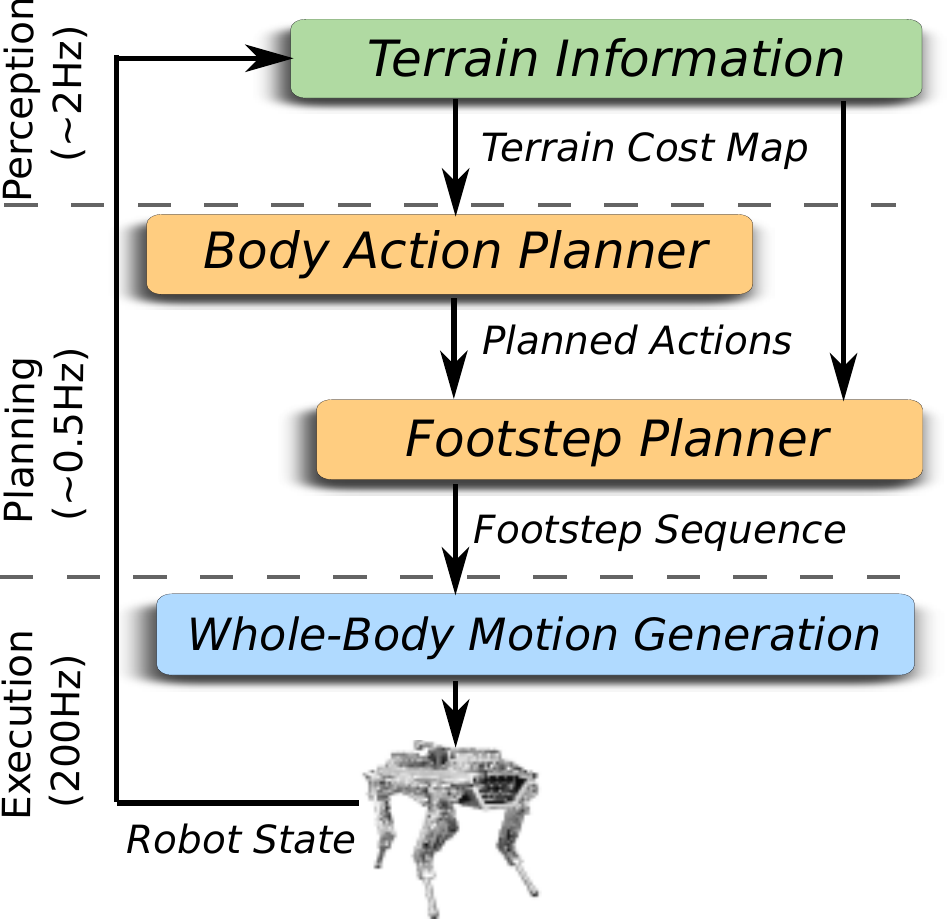}
	\caption{An overview of the perception and planning system that generates
	footstep sequences according to the terrain information.}
	\label{fig:framework}
	\vspace{-10pt}
\end{figure}

This section describes the pipeline from the acquisition and evaluation of
terrain information to the generation of appropriate footholds
(\fref{fig:framework})\footnote{A more in-depth presentation of the perception and terrain evaluation pipeline can be found in \cite{tepra15planning}.}. The on-board terrain information server continuously
holds the state of the environment. The body action planner decides the general
direction of movement and the footstep sequence planner chooses specific
footholds along this path.

\subsection{Terrain Information}
We develop a terrain information server that computes the required
information for the body action and the footstep sequence planners, e.g. the
\textit{terrain cost map} of the environment. We build a 3D occupancy grid map
\cite{hornung13auro} from a RGBD sensor mounted on a scanning pan \& tilt unit,
alongside with the state estimate of the robot using the Extended Kalman Filter
\cite{bloesch12state}. The voxel-based map is built using a $(x,y,z)$ resolution
of ($\unit[4]{cm} \times \unit[4]{cm} \times \unit[2]{cm}$) which roughly matches
 the dimensions of the robot's foot. 

The terrain cost map quantifies how desirable it is to place a foot at a
specific location. The terrain cost $c_t$ for each voxel in the map is computed
using geometric terrain features as in \cite{winkler2014}. Namely, we use
the standard deviation of height values, the slope and the curvature of the cell
in question. The terrain cost $c_t$ for each voxel of the map is computed as a
weighted linear combination of the individual features $c_t(x,y) = \mathbf{w}^T
\mathbf{c}(x,y)$. 
The cost map is
locally re-computed (in a \unit[2.5]{m}$\times$\unit[5]{m} area around the
robot) whenever a change in the map is detected.

\newcommand{\smb}{\mathbf{s}}

\subsection{Body Action Planner}\label{sec:body_action_planner_alex}
The state ${\mathbf{s}=(x,y,\theta)}\in\mathcal{S}$ of the robot body includes
the current position $(x,y)$ and the yaw angle $\theta$. Given a desired goal state,
the body action planner finds a sequence of actions $\mathbf{a}_{0\dots N} =
\{\mathbf{a}_0, \mathbf{a}_1,\dots, \mathbf{a}_N\}$ that move the robot in a
nearly optimal way to this state. This implies that terrain features, the
difficulty of specific actions, kinematic reachability and collision with the
environment must be considered and quantified. A feasible action $\mathbf{a}$ is 
the change of state that can be achieved through one step 
from $\mathbf{s}$ to $\mathbf{s'}$ as
\begin{equation}
	\renewcommand{\dx}{\Delta x}
	\newcommand{\dy}{\Delta y}
	\newcommand{\dth}{\Delta \theta}
	\mathbf{a} = (\dx, \dy, \dth ) \in \mathcal{A}.
\end{equation}

We define $\mathcal{A}$ as the set of -empirically chosen- feasible
motion primitives (e.g. move left, diagonally forward, back) that correspond to the
kinematics and dynamics of the robot. The cost of an action $\mathbf{a}$ given a
current state $\mathbf{s}$ is computed as a weighted linear
combination of costs:
\begin{equation}
c(\mathbf{s},\mathbf{a}) = \mathbf{w}^T \mathbf{c}(\mathbf{s},\mathbf{a})
\end{equation}
with $\mathbf{c}(\mathbf{s},\mathbf{a})$ consisting of:
\begin{description}
	\item [$\bar{c}_{t}$] The average of the best n terrain costs around each leg 
	after performing action $\mathbf{a}$.
	\item [$c_{a}$] The difficulty of a specific action, e.g. sideways steps 
	are 	more difficult than forward ones.
	\item [$c_{pc}$] Penalizes actions that potentially cause the swing-leg to 
	collide with the environment.
	\item [$c_{po}$] Penalizes actions that potentially end up in uneven terrain
	that require large roll and pitch angles.
\end{description}

\begin{figure}%
	\centering
	\includegraphics[width=0.98\columnwidth]{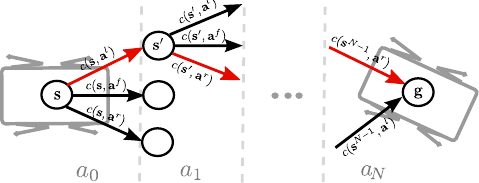}
	\caption{A sketch of the body action graph. The objective is to find
	a sequence of actions
	$\mathbf{a}$ from the current body state ${\mathbf{s}=(x,y,\theta)}$ to the
	goal state $\mathbf{g}$, that minimizes the accumulated action costs
	$c(\mathbf{s},\mathbf{a})$. For simplicity only three possible actions are
	shown, namely move left ($\mathbf{a}^l$), right ($\mathbf{a}^r$) and forward
	($\mathbf{a}^f$). The optimal action sequence \{$\mathbf{a}^l, \mathbf{a}^r,
	\dots, \mathbf{a}^r$\} found through ARA* is shown in red.}
	\label{fig:body_action_graph}
	\vspace{-10pt}
\end{figure}

The set of actions $\mathcal{A}$ and the current state $\mathbf{s}$ of the
robot is used to construct a directed graph $\mathcal{G} = (\mathcal{S},\mathcal{A})$ (\fref{fig:body_action_graph}).
We use the ARA* \cite{likhachev2004} algorithm to search the tree for a sequence
of actions with the lowest accumulated cost from the current to the goal state.
ARA* uses a heuristic $h(\mathbf{s})=%
-\bar{c}\mathcal{F}(\|\mathbf{g}-\mathbf{s}\|)$ to decide along which states to
search first. $\mathcal{F}(\cdot)$ is the estimated remaining steps
(actions) to reach the goal state and $\bar{c}$ is an estimated lower bound
on the average future action costs, considering the terrain costs between the
current and the goal state.

ARA* initially runs an A* search with an inflated heuristic, ${\epsilon \cdot
h(s)}$, which quickly finds a first sequence of actions. Unfortunately, since
the inflated heuristic is no longer admissible (always lower than the true
cost), the sequence of actions may be sub-optimal. As long as computational time
is still available, ARA* repeatedly runs A* search, continuously decreasing the
inflation factor $\epsilon$ and thereby finding closer to optimal sequences of
actions. Since a first solution, although suboptimal, is found quickly, this
algorithm can be used online.
  
\subsection{Footstep Sequence Planner}
Given the desired body action plan, the footstep sequence planner computes the
sequence of footholds that corresponds to these body actions. In our previous
work, we selected the optimal foothold around the nominal stance positions in a
predefined swing-leg sequence. In this paper, we modify the position of the
search area and the swing-leg sequence depending on the corresponding action,
which improves the robustness of the planned actions. For example, when moving
left (action $\mathbf{a}^l$) it is advantageous to swing one of the left legs to 
avoid small areas of support.

The footstep location in each search area with the lowest foothold cost, $c_f=
w_t c_t + w_{st} c_{st} + w_c c_c + w_o c_o$, is then selected, where $c_t$ is
the terrain cost below a foothold, $c_{st}$ is the support triangle cost,
$c_c$ is the leg collision cost and $c_o$ is the body orientation cost.

\begin{figure}[!b]
	\centering
	\includegraphics[width=1.0\columnwidth]{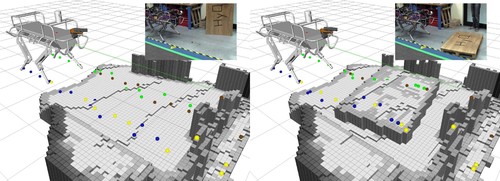}
	\caption{(Re-)planning and perception. The left image shows how a map of the
	environment is built (cost values in grayscale) along with the body path
	(green line) and the footstep
	sequence plan (colored spheres). Once the environments changes the map
	is updated and the footsteps re-planned.}
	\label{fig:pallet_replan}
\end{figure}

\subsection{Re-planning/Updating during Execution}
Compared to our previous approach the graph is significantly smaller,
since we only
search over feasible actions $\mathcal{A}$ and not over every discretized
change in state. Additionally, ARA* provides intermediate solutions, so the
exhaustive and time-costly search procedure does not need to be completed before
the robot can react. This combination of the efficient voxel-based occupancy
map, the graph representation over feasible actions, and the efficient search
through ARA* allows us to re-plan the motions and set of planned footholds
online to cope with changes in the environment as is shown in Fig.
\ref{fig:pallet_replan}.

\section{Whole-Body Motion Generation}
\label{sec:trajectory_generation}
We generate a body trajectory that ensures that the robot is \emph{dynamically 
stable} at every time step. 
We follow the approach presented in \cite{kalakrishnan2010IJRR}
that finds a \gls{cog} trajectory that respects stability
constraints without explicitly generating a \gls{zmp} trajectory. We build
on this approach by enabling swing-leg sequences in any order through insertion of four-leg support phases.

\subsection{Problem Formulation}
\label{subsec:zmpOptimization}
For a \gls{cog} trajectory to be feasible it must be continuous and double 
differentiable. This way we avoid steps in accelerations that produce discontinuous 
torques which can damage the hardware and affect stability. 
The body trajectory, $x_{cog}$, is given by a spline %
comprised of multiple fifth-order polynomials:
\begin{equation}
  x_{cog}(t) = a_xt^5 + b_xt^4 + c_xt^3 + d_xt^2 + e_xt + f_x.
  \label{eq:poly}
\end{equation}
At each spline junction we require the
last state ($t=T_i$) of spline $i$ to be equal to the first state ($t=0$) of the
next spline $i\!+\!1$ as:
\begin{equation}
  (x_{cog}, \dot{x}_{cog}, \ddot{x}_{cog})_{t=T_i}^i = (x_{cog}, \dot{x}_{cog},
  \ddot{x}_{cog})_{t=0}^{i+1}.
  \label{eq:equality}
\end{equation}
This ensures double differentiability and continuity of the trajectory, required by the 
floating-base inverse dynamics. Finding an optimal \gls{cog} trajectory can then be reduced to finding optimal polynomial coefficients
${\bvec{q}_i}~=~(a_x, \ldots, f_x, a_y, \ldots, f_y)^T \in \mathbb{R}^{12}$ for each spline segment $i$.

\subsection{Dynamic Stability}
\begin{figure}[tb]
  \centering
  \subfigure{\resizebox{0.45\columnwidth}{!}{
  \includegraphics{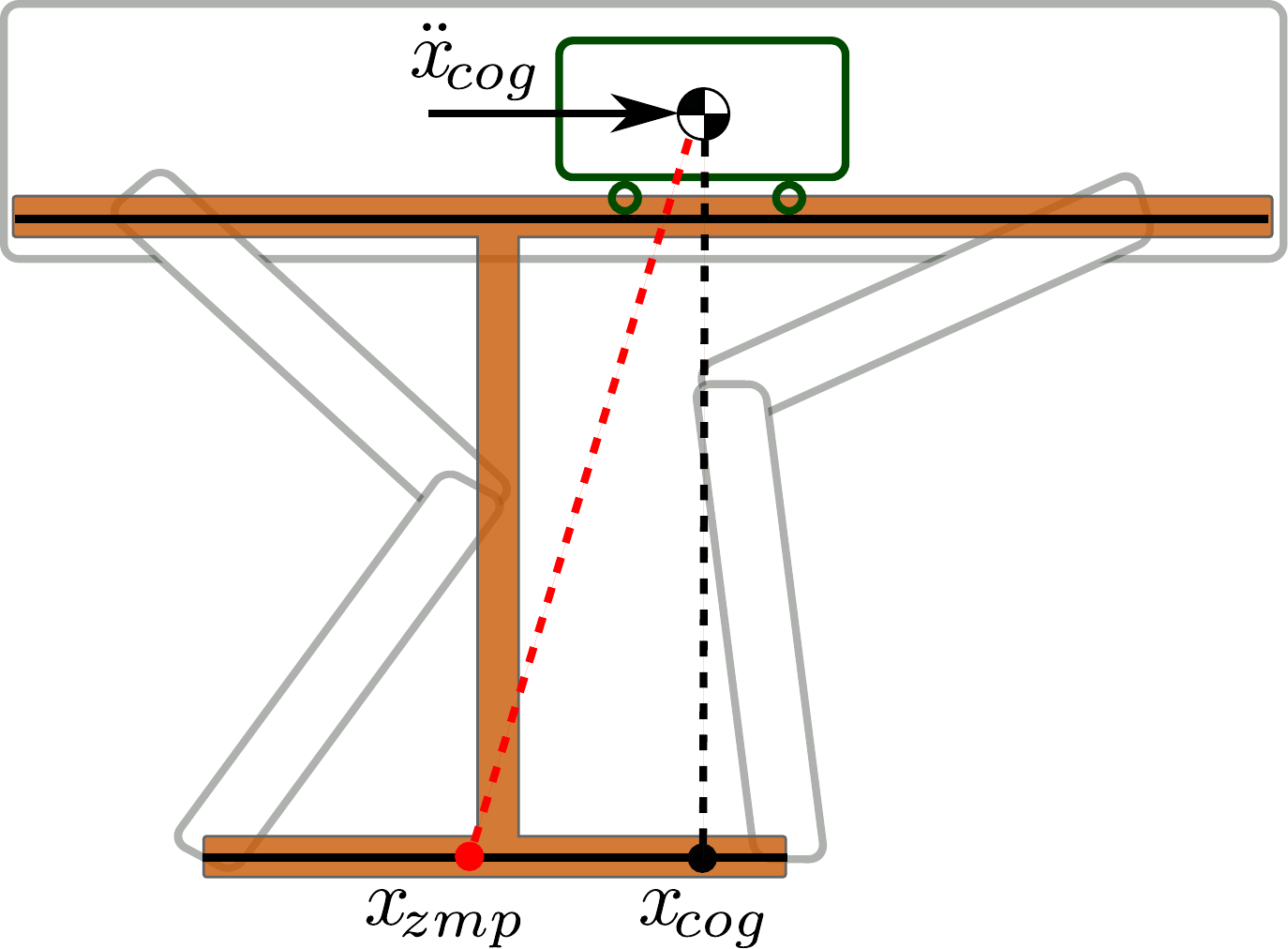}}}
  \subfigure{\resizebox{0.5\columnwidth}{!}{
  \includegraphics{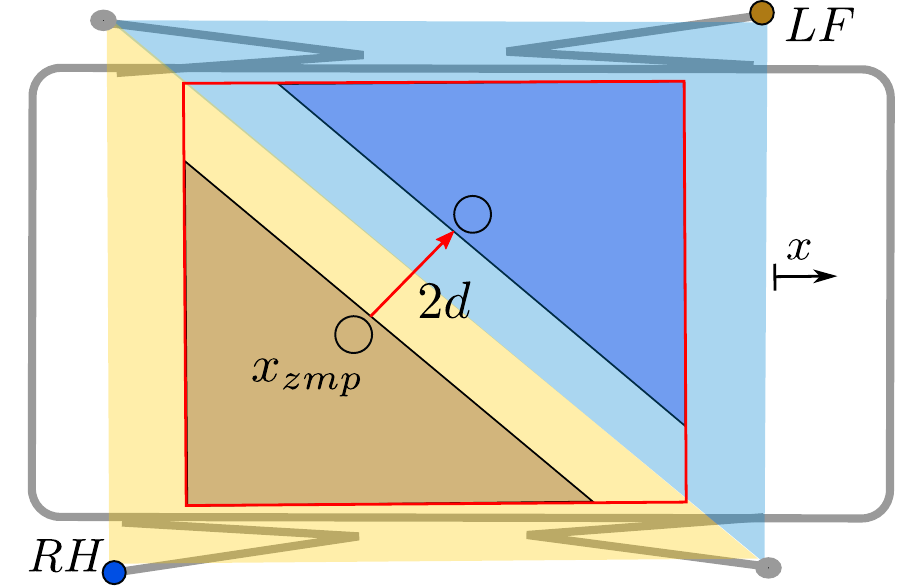}}}		
  \caption{\textit{Left:} Cart-Table model for representing a quadruped robot: The total mass of the robot is concentrated in the cart that moves on the table.
  The base of the table represents the current area of support, determined by the current footholds. The \gls{zmp} must lie inside this area for dynamic stability.
  \textit{Right:} Disjoint support triangles due to the added stability margin $d$.
  When switching between swinging the left-front (LF) to right-hind (RH)
  the ZMP must move from the brown to the blue support triangle. Since all four feet are in stance during this phase, the ZMP is only restricted by the red support polygon.}
  \label{fig:supp_tr}
  \vspace{-10pt}
\end{figure}

To execute the planned footsteps, a body trajectory must be found that ensures 
a stable stance at all time instances. For slow movements this is achieved by keeping the \gls{cog} inside the support polygon, i.e. the polygon formed by the legs in stance. To consider \emph{dynamic} effects of larger body accelerations we estimate the position of the \gls{zmp} by modeling the robot as a cart-table (\fref{fig:supp_tr}, left). The \gls{zmp} can then be calculated by:
\begin{equation}
  x_{zmp} = x_{cog} - \frac{z_{cog} \ddot{x}_{cog}} {\ddot{z}_{cog} + g_0},  
  \label{eq:zmp}
\end{equation}
where $x_{zmp}$ and $x_{cog}$ are the position of the \gls{zmp} and the \gls{cog}
respectively, $z_{cog}$ describes the height of the robot with respect to its feet, $\ddot{z}_{cog}$ is the vertical acceleration of the body and $g_0$
represents the gravitational acceleration. Dynamic stability requires the \gls{zmp} to be inside the current support triangle, expressed by three lines $l$ of
the form $px + qy + r = 0$. The \gls{zmp} is considered to be \emph{inside} a
support triangle, if the following conditions are met at every sampling interval: 
\begin{equation}
  p_l x_{zmp} + q_l y_{zmp} + r_l > 0 \quad \text{ for } l=1,2,3.
  \label{eq:line}
\end{equation}

In reality there exist discrepancies between the cart-table model and the real robot. Additionally, desired body trajectories cannot be perfectly tracked as modelling, 
sensing and actuation inaccuracies are hard to avoid. Therefore, it is best to avoid
the border of stable configurations by shrinking the support triangles
by a stability margin~$d$ (\fref{fig:supp_tr}, right). 
With the introduction of $d$, there is no 
continuous \gls{zmp} trajectory when switching between diagonally opposite 
swing legs as the support triangles are disjoint. We therefore allow a transition period (`\textit{four-leg support phase}') during which the \gls{zmp} is only restricted by the shrunk support \emph{polygon} created by the four stance feet.

We built on \cite{kalakrishnan2010IJRR}
by allowing a \emph{completely irregular sequence} of steps for the 
\gls{zmp} optimization. Our trajectory generator needs no knowledge of a
predefined gait. For \emph{every} step it checks
if the next swing leg is diagonally opposite of the
current swing leg. If so, the disjoint support triangles require a four-leg support phase for the optimization to find a solution. This allows a greater
decoupling from the footstep planner, which can generate swing leg sequences in
any order useful for the success of the behaviour.

\subsection{Cost function}
In addition to moving in a dynamically stable way, the trajectory should accelerate as little as possible during the execution period $T$. This increases possible execution speed and reduces required 
joint torques. This is achieved by minimizing
\begin{equation}
  J = w_x \int_{0}^{T}\! \ddot{x}_{cog}^2(t)\,dt + w_y\int_{0}^{T}\!
  \ddot{y}_{cog}^2(t)\,dt.
  \label{eq:cost_fun}
\end{equation}
The directional weights \bvec{w} penalize sideways accelerations ($w_y =
1.5w_x$) more than forward-backward motions, since sideways motions are more
likely to cause instability. This results in a convex quadratic program (QP) with the cost function~(\ref{eq:cost_fun}), the equality constraints~(\ref{eq:equality}) and the inequality constraints~(\ref{eq:line}). We solve it using the freely available QP solver, namely \textit{QuadProg++} \cite{guennebaud2011} to obtain the spline coefficients~$\mathbf{q}$ and therefore the desired and stable $(x,y)$-body trajectory~(\ref{eq:poly}). The remaining degrees of freedom ($z_{cog}$, roll, pitch, yaw) and the swing-leg trajectories are chosen based mostly on the foothold heights and described in detail in \cite{winkler2014}.

\section{Execution of Whole-Body Motions}\label{sec:execution}
\begin{figure}[tb]
  \centering  
\subfigure{\resizebox{0.75\columnwidth}{!}{
\includegraphics{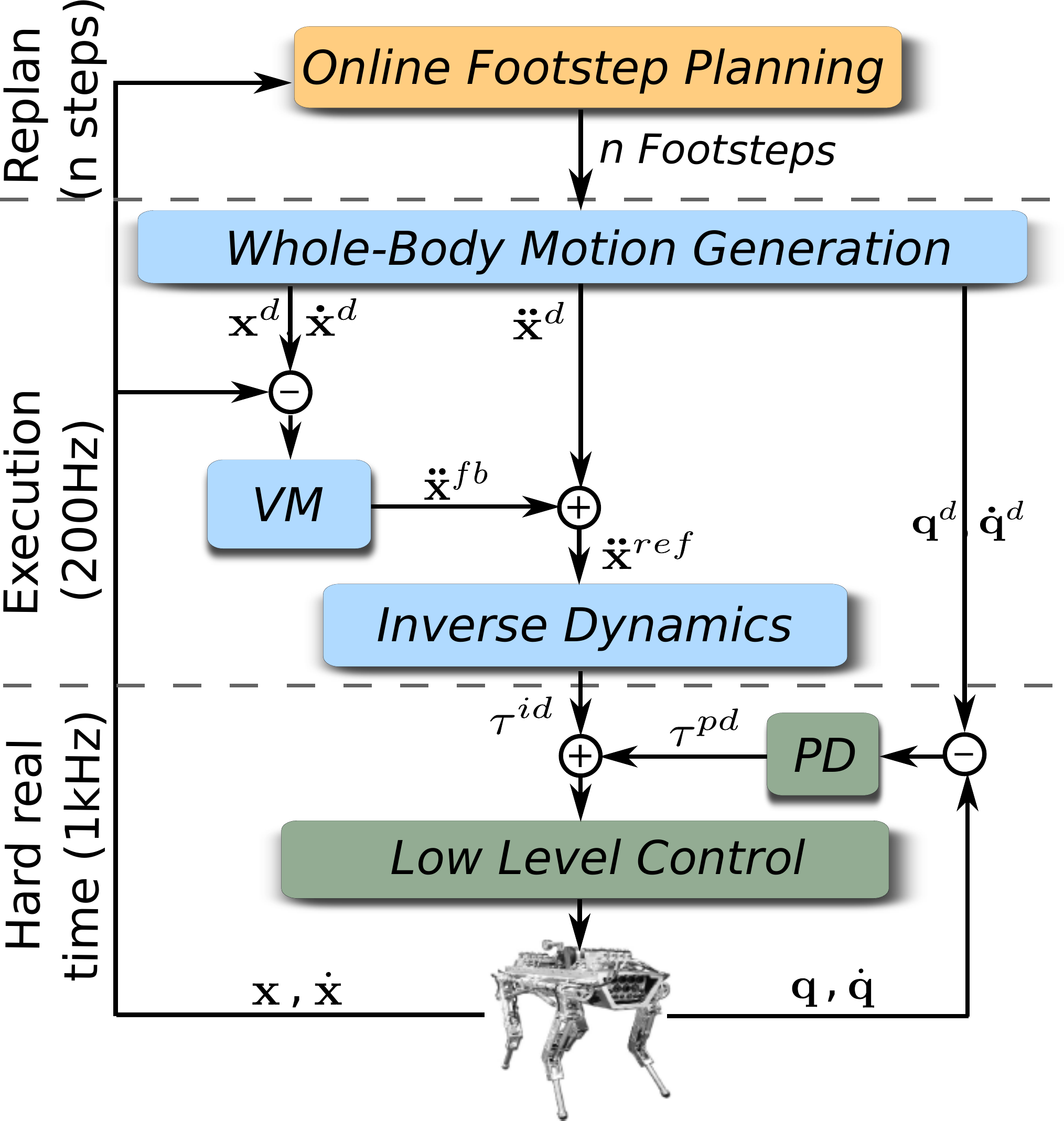}}}
  \caption{Pipeline that uses planned footholds to generate dynamic whole-body 
	motions and compliantly executes them using a combination of feedforward and
	feedback terms.} 
  \label{fig:controller}
  \vspace{-10pt}
\end{figure}

Dynamic whole-body motions require orchestrated and precise actuation
of all the joints. Simple PD controllers do not suffice for such
motions, especially when considering uncertainties in the environment 
and/or model inaccuracies. 
We use a control scheme (\fref{fig:controller}) that combines a virtual 
model with a floating-base inverse dynamics controller. 
After receiving an arbitrary sequence of footholds from the footstep planner, the
whole-body motion generator calculates desired (feedforward) accelerations
$\vc{\ddot{x}}^d$ for the body and a virtual model (VM) control loop adds feedback
accelerations $\vc{\ddot{x}}^{\!f\!b}$ should the robot deviate from the desired
trajectory. The inverse dynamics produce the majority of the joint torques which 
are combined with a low-gain joint-space PD controller to
compensate for possible model inaccuracies. The computed reference torques are 
then tracked by the low-level torque controller. Note that $\vc{x}$ describes the
linear and rotational coordinates of the body as
\begin{equation}
 \vc{x} = (\vc{x}_{cog}, \mx{R}_{b}),
 \vc{\dot{x}} = (\vc{\dot{x}}_{cog}, \vc{\omega}_b),
 \vc{\ddot{x}} = (\vc{\ddot{x}}_{cog}, \vc{\dot{\omega}}_b), 
\end{equation}
where $\mx{R}_b \in \Rnum^{3\times 3}$ is a coordinate rotation matrix representing 
the orientation of the base w.r.t. the world frame and 
$\vc{\omega}_b \in \Rnum^3$ is the angular velocity of the base.

\subsection{Virtual Model}
The feedback action which compensates for inaccurate execution and drift
can be imagined as virtual springs and dampers attached to the robot's trunk on 
one side and the desired body trajectory on the other \cite{pratt2001}. 
Deviation between these causes the springs and dampers to produce virtual 
forces $\vc{F}_{vm}$ and torques $\vc{T}_{vm}$ on the body that ``pull'' the 
robot back into the desired state through 
\begin{equation}
\begin{aligned}
	& \vc{F}_{vm} = \mx{P}_x (\vc{x}_{cog}^d - \vc{x}_{cog}) + \mx{D}_x (\vc{\dot{x}}_{cog}^d - \vc{\dot{x}}_{cog})\\
	& \vc{T}_{vm} =  \mx{P}_{\theta} e(\mx{R}_b^d \mx{R}_b^\T) + \mx{D}_{\theta} (\vc{\omega}_b^d - \vc{\omega}_b),  
\end{aligned}
\label{eq:virtual_model}
\end{equation}
where the superscript $d$ refers to the desired values, planned by the 
whole-body motion generator and non-superscript values describe the state 
of the robot as estimated by the on-board state estimator. 
Respectively, $e(.): \Rnum^{3\times 3} \rightarrow \Rnum^3$ is a mapping 
from a rotation matrix to the associated rotation vector \cite{Caccavale1999}. 
$\mx{P}_x, \mx{D}_x,\mx{P}_{\theta}, \mx{D}_{\theta}\in \Rnum^{3\times 3}$ 
are positive-definite diagonal matrices of proportional and derivative gains, 
respectively. Expressing the body feedback action in terms of forces and 
moments allows us give the virtual model gains a physical meaning of 
stiffness and damping and thus can be intuitively set and used. 

Since the inverse dynamics computation requires reference \textit{accelerations}, 
we multiply the forces/moments (\textit{wrench}) $\vc{\mathcal{W}}_{vm} = ( \vc{F}_ {vm}, \vc{T}_{vm})$ with the
inverse of the composite rigid body inertia $\mx{I}_c$ of the robot
at its current configuration. Adding this body 
feedback acceleration to the desired body acceleration produced by the 
whole-body motion generator creates the 6D reference acceleration 
(linear and rotational) for the inverse dynamics computation as:
\begin{equation}
	 \vc{\ddot{x}}^{ref}= \vc{\ddot{x}}^d  + I_c^{-1}\vc{\mathcal{W}}_{vm} .
\label{eq:desired_acc}
\end{equation}
By combining a feedforward acceleration $\vc{\ddot{x}}^d$ with a 
body-feedback acceleration, we achieve accurate tracking while 
maintaining a compliant behaviour.  

\subsection{Floating Base Inverse Dynamics}
The floating base inverse dynamics algorithm calculates the joint 
torques required to execute the reference body accelerations. We can
partition \cite{Fujimoto1998} the dynamics equation of the robot 
into the unactuated base coordinates $q_b \in \Rnum^6$ and the 
active joints' $q \in \Rnum^{12}$ as
\begin{equation}
	\underbrace{\mx{M}(\vc{R}, \vc{q})  \mat{\vc{\ddot{q}}_b \\ \vc{\ddot{q}}} + 
	\mat{ \vc{h}_b \\ \vc{h}_q }(\vc{R}, \vc{q}, \vc{\omega}, \vc{\dot{q}})}_\vc{b}   = 
	\mat{	\vc{0}\\ \vc{\tau}} + \mat{\mx{J}_{cb}^T\\ \mx{J}_{cq}^T } \vc{\lambda},
\label{eq:invDyn}
\end{equation}
where $\mx{M}$ is the floating base mass matrix, $\vc{h} = (\vc{h}_{b},  \vc{h}_{q})$
is the force vector that accounts for Coriolis, centrifugal, and gravitational forces,
$\vc{\lambda}$ are the ground contact forces, 
and their corresponding Jacobian $\mx{J}_c = \mat{ \mx{J}_{cb} &  \mx{J}_{cq}}$ and
$\vc{\tau}$ are the torques that we wish to calculate.

The left hand term $\vc{b} = (\vc{b}_{b},  \vc{b}_{q})$ can be 
computed efficiently using the Featherstone implementation of the Recursive 
Newton-Euler Algorithm (RNEA) \cite{Featherstone2008}. Since the \gls{cog} acceleration 
$\vc{\ddot{x}}_{cog}^{ref}$ is defined in a 
frame aligned with the base frame but with the origin in the \gls{cog}, we 
perform a translational coordinate transform $_b\mx{X}_{cog}$ to get the 6D base spatial acceleration:  
$\vc{\ddot{q}}_b  =\,  _b\mx{X}_{cog} \vc{\ddot{x}}^{ref}$ as in \cite{Featherstone2008}.

By partitioning the dynamics equation as in \eqref{eq:invDyn} and given that the
base is not actuated, we can directly compute, in a least-squares way, the 
vector of ground reaction forces $\vc{\lambda}$ from the first $n_b$ equations, 
$\vc{\lambda} = \vc{J}_{cb}^+ \vc{b}_b$, where $()^+$ denotes the Moore-Penrose 
generalized inverse. We then use the last $n$ equations to produce the 
reference joint torques, $\vc{\tau}^{id} = \vc{b}_q- \mx{J}_{cq}^T\vc{\lambda}$.  

\section{Experimental Results}\label{sec:experimental_results}
This section describes the experiments conducted to validate the 
effectiveness and quantify the performance of our framework.

\subsection{Experimental Setup}
We use the hydraulically-actuated quadruped robot 
HyQ in our experiments.
HyQ weighs approximately \unit[90]{kg}, is
fully-torque controlled and equipped with precision joint encoders, 
a depth camera (Asus Xtion) and an Inertial Measurement Unit (MicroStrain). We perform on-board state
estimation and do not make use of any external state measuring system, e.g. a 
marker-based tracking system. All computations are done on-board, using a
PC104 stack for the real-time critical part of the framework, and a 
commodity i7/\unit[2.8]{GHz} PC for perception and planning.

The first experiment starts with flat, obstacle-free terrain. After the robot has
planned initial footsteps, a pallet is placed into the terrain. 
In the next experiments the robot must climb one and two pallets of dimensions
\unit[1.2]{m}~$\times$~\unit[0.8]{m}~$\times$~\unit[0.15]{m}. The height of 
one pallet is equal to 20\% of the leg length.
Furthermore we show that the robot traverses a gap of \unit[35]{cm}, which is 
approximately half the robot's body length.
The final experiment consist of two pallets connected by a sparse path of
stepping stones. The pallets are \unit[1.2]{m} apart and the stepping stones
lie \unit[0.08]{m} lower than the pallets.

\begin{figure*}[tb]
  \centering
\subfigure{\resizebox{1.0\textwidth}{!}{
\includegraphics{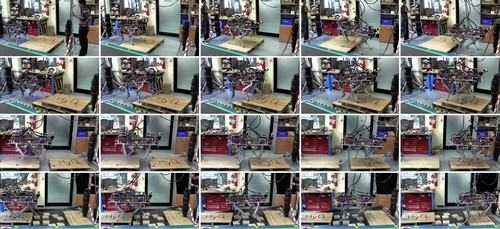}}}
  \caption{Snapshots of experimental trials used to evaluate the performance
  of our framework. From top to bottom: crossing a \unit[15]{cm} pallet; climbing a stair-like structure consisting of two stacked pallets; traversing a \unit[35]{cm} gap and crossing a sparse set of stepping stones.
  }
  \label{fig:vidsnaps}
  \vspace{-5pt}
\end{figure*}

For each experiment, we specify the $(x,y, \theta)$ goal state. The footstep planner finds a sequence of footsteps of an arbitrary order, 
which the controller then executes dynamically. We validate the performance of 
our framework in 4 scenarios as seen in \fref{fig:vidsnaps} and compare it to 
our previously achieved results (\tref{tab:results}) on the same benchmark 
tasks. Additionally, the reader is strongly encouraged to view the
\href{http://youtu.be/MF-qxA_syZg}{accompanying 
video}\footnote{\url{http://youtu.be/MF-qxA_syZg}}
as it provides the most intuitive way to demonstrate the performance of our
framework.

\subsection{Results and Discussion} \label{sec:evaluation}
\begin {table}[!t]
\caption{Forward speed and success rate of experiments averaged over 10 trials and compared to previous results from \cite{winkler2014}.}
 \label{tab:results} 
\begin{center}
\begin{tabular}{@{} lrrrrrrr @{}}    
\toprule 
&\multicolumn{3}{c}{Speed [cm/s] } && \multicolumn{3}{c}{Success Rate [\%]} \\
\cmidrule{2-4}\cmidrule{6-8}
\emph{Terrain} 	& Curr. &Prev. &Ratio &&   Curr. &Prev. &Ratio\\
\midrule
 Step. Stones  	&  7.3  &1.7   &4.2  &&   60    &70  &0.9      \\ 
 Pallet 	    &  9.5  &2.1  &4.5   &&   100   &90  &1.1       \\ 
 Two Pallets  	& 10.2  &1.8  &5.8   &&   90    &80  &1.1       \\
 Gap			& 12.7  &\enspace- &\enspace- &&   90   &\,\,0   &\enspace-    \\
 \bottomrule
\end{tabular}
\end{center}
\vspace{-15pt}
\end {table}  
\subsubsection{Perception and (re-)planning}
Efficient occupancy grid-based mapping and focusing our computations to an area
of interest around the robot body greatly increase computation speed. This allows
us to incrementally build a model of the environment and update the terrain cost 
map at a frequency of \unit[2]{Hz}. Using the action based search graph together with
ARA* allows us to replan footholds at a frequency of approximately
\unit[0.5]{Hz} for goal states up to \unit[5]{m}. 

\subsubsection{Speed while dynamically stable} 
The pallet climbing and gap experiments show the \emph{speed} (\tref{tab:results}) that
our framework can achieve: 
This is due to the fact, that the body can move
faster while still being stable, since we are using a \emph{dynamic} stability
criterion. All accelerations and decelerations are optimized, so that the
\gls{zmp} never leaves the support polygon. 
In addition, since we are not directly producing torques with the virtual model
feedback controller, but only accelerations for the inverse dynamics controller,
our feedback actions also respect the dynamics of the system. 
Furthermore, the
duration of the four-leg-support phase is significantly reduced: It is much
faster to move the \gls{zmp} from one support triangle to another than the
\gls{cog} (e.g. entire body), because this can be achieved by manipulating
the acceleration. 
\subsubsection{Model accuracy} 
Walking over a \unit[35]{cm} gap (approximately half of the body length) shows the stability of the robot despite of highly dynamic motions. When crossing the gap the robot accelerates up to a body velocity of \unit[0.5]{m/s} 
and is able to decelerate again without loosing balance. This shows, that the
simple cart-table model is a sufficient approximation for large quadrupeds performing locomotion tasks.
\subsubsection{Avoiding kinematic limits} 
Attempting to cross the gap with a statically stable gait tends to overextend 
the legs, since large body motions are required to move the robot into statically
stable positions. Dynamic motions allow us to keep the \gls{cog} closer to the 
center of all four feet, since stability can be achieved by appropriate
accelerations, avoiding kinematic limits.
\begin{figure}[tb]
	\centering
	\subfigure{\resizebox{0.98\columnwidth}{!}						
	{\includegraphics{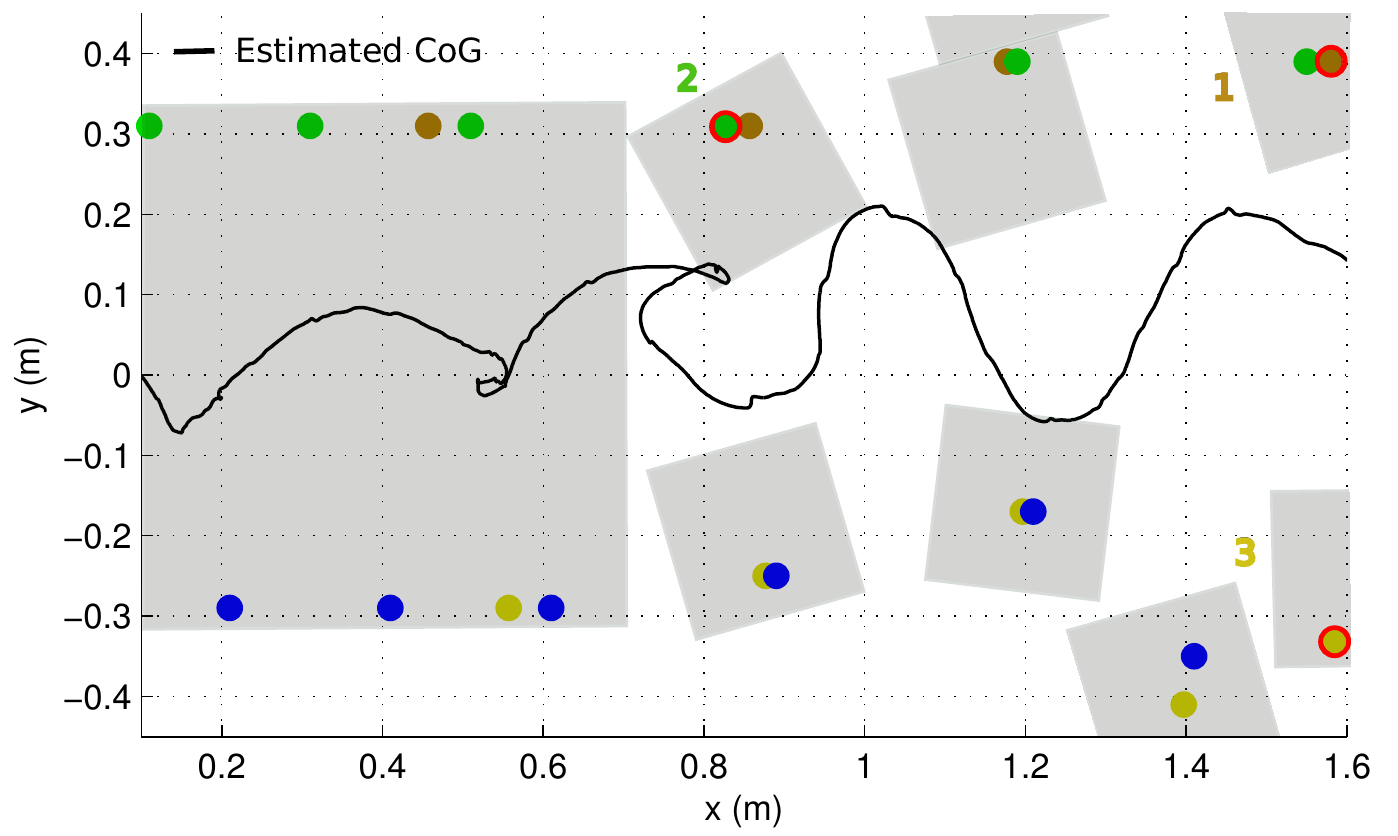}}
	\label{fig:cog_trajectory_real}}
	\subfigure{\resizebox{0.98\columnwidth}{!}						
	{\includegraphics{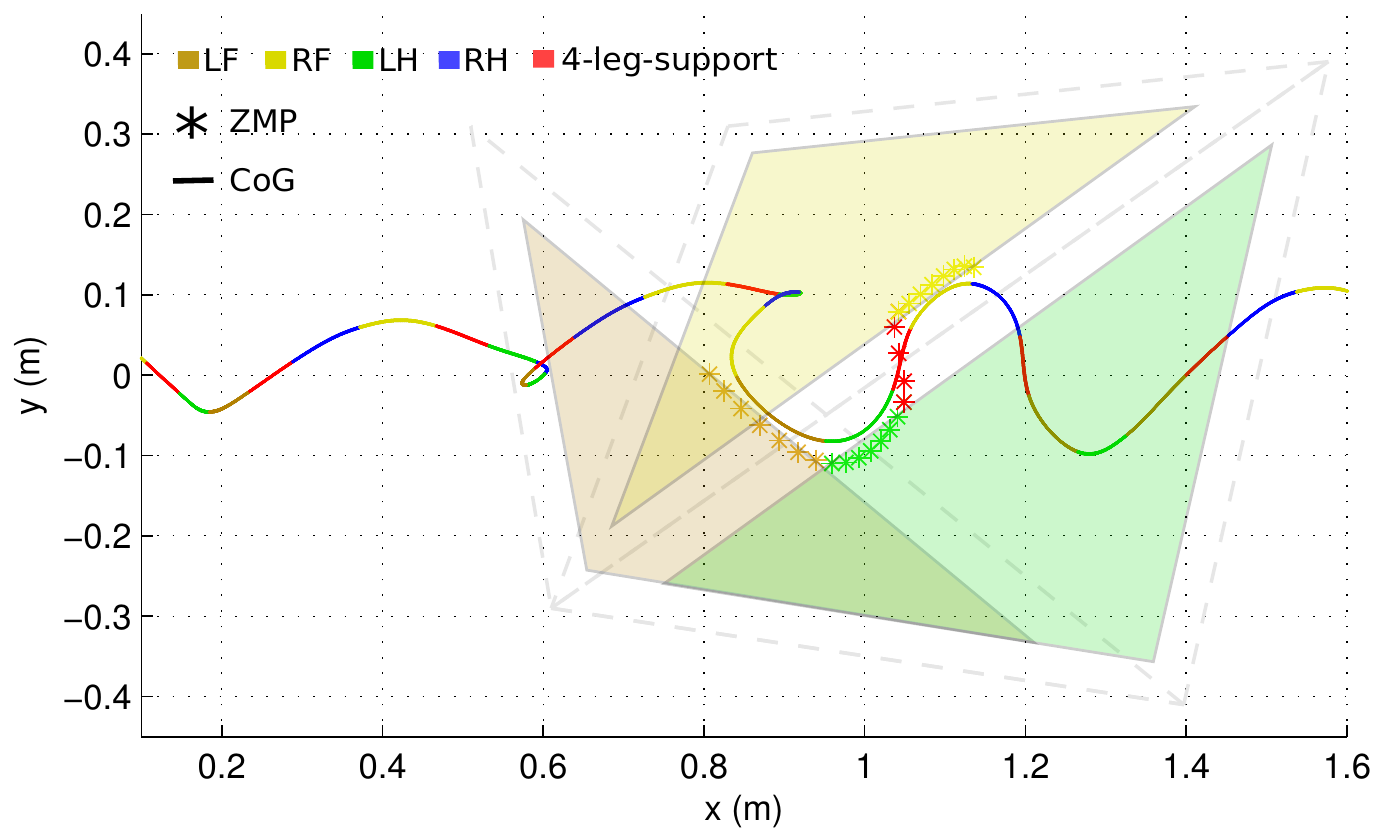}}
	\label{fig:cog_trajectory_des}}
	\caption{ %
	\textit{Top}: The body motion when walking over the stepping stones is show in black. The planned footholds are shown and the irregular step sequence LF(1)~$\rightarrow$~LH(2)~$\rightarrow$~RF(3) is highlighted (red). 
	\textit{Bottom}: The 3 shrunk support triangles corresponding to the highlighted step sequence brown~$\rightarrow$~green~$\rightarrow$~yellow are shown. Additionally the planned CoG (solid line) and ZMP trajectory for the duration of these 3 steps is illustrated (asterix). While the CoG (solid line) does not reach the support triangles, the ZMP does, causing dynamic stability. When switching between disjoint support triangles (green~$\rightarrow$~yellow) four-leg support phases are inserted (red) to allow a smooth transition.}  
	\label{fig:cog_trajectory}
	\vspace{-10pt}
\end{figure} 
\subsubsection{Stability despite irregular swing-leg sequences} 
Walking over the stepping stones demonstrates the ability of the controller to execute irregular swing-leg sequences in a dynamically stable manner (\fref{fig:cog_trajectory}).
Starting from a lateral sequence gait (LH-LF-RH-RF) the foothold sequence changes to traverse these irregularly placed stepping stones. Despite this, the produced \gls{cog} trajectory (colored solid line) is dynamically stable, since the \gls{zmp} (asterisk) is always kept inside the current support triangle. When comparing the actual (top) and desired (bottom) CoG trajectories, a tracking error is evident. By keeping the \gls{zmp} e.g. d = \unit[6]{cm} away from the stability borders, we are robust even against these inaccuracies. 

The whole body motion generator inserts four-leg-support phases (red section) only whenever it detects disjoint support triangles in the swing-leg sequence. While executing steps 1 and 2 (\fref{fig:cog_trajectory}) no four-leg-support phase is necessary, because the triangles are not disjoint. Only after returning to swing the right-front leg, the robot requires a four-leg support phase for the \gls{zmp} to transition from the green (LH) to the yellow (RF) support triangle at $(x,y)=(1.1, 0)$.

\section{Conclusion}\label{sec:conclusion}
We presented a dynamic, whole-body locomotion framework that executes
footholds planned on-board. We showed, how a change in the environment causes
the foothold generator to re-plan footholds on-line. We presented a whole body 
motion planner, which is able to generate a ZMP-stable body trajectory despite irregular swing-leg sequences to execute footholds dynamically. 
We showed how a combination of virtual model and floating-base 
inverse dynamics control can compliantly, yet accurately,
track the desired whole-body motions. Real world experimental trials on challenging terrain demonstrate the capability of our framework.  

We are currently working on bringing the kinematic planning and dynamic execution closer together. The idea is to produce desired state trajectories and required torques through \emph{one} trajectory optimization problem, taking into account torque/joint limits, the dynamic model of the robot, foothold positions, friction coefficients and other constraints. With this approach we aim to produce even more dynamic motions such as jumping and rearing, during which fewer or no legs are in contact.

\section*{Acknowledgements} \footnotesize \vspace{-2pt} This research is funded
by the Fondazione Istituto Italiano di Tecnologia. Alexander W. Winkler was partially supported by the Swiss National Science Foundation (SNF) through a Professorship Award to Jonas Buchli and the NCCR Robotics.
The authors would like to thank the colleagues that collaborated for the success
of this project: Roy Featherstone, Marco Frigerio, Marco Camurri, Bilal Rehman, Hamza Khan, Jake Goldsmith, Victor Barasuol, Jesus Ortiz, Stephane Bazeille and our team of technicians.

\bibliographystyle{./IEEEtran}
\bibliography{IEEEabrv,all_in_one.bib}
\end{document}